\documentclass[sigconf,authorversion]{acmart}

\copyrightyear{2023}
\acmYear{2023}
\setcopyright{acmlicensed}\acmConference[ICPP 2023]{52nd International Conference on Parallel Processing}{August 7--10, 2023}{Salt Lake City, UT, USA}
\acmBooktitle{52nd International Conference on Parallel Processing (ICPP 2023), August 7--10, 2023, Salt Lake City, UT, USA}
\acmPrice{15.00}
\acmDOI{10.1145/3605573.3605609}
\acmISBN{979-8-4007-0843-5/23/08}

\usepackage{graphicx}
\usepackage{textcomp}
\usepackage[linesnumbered,ruled]{algorithm2e}
\usepackage{tabu,booktabs}

\usepackage{subcaption}

\AtBeginDocument{%
  \providecommand\BibTeX{{%
    \normalfont B\kern-0.5em{\scshape i\kern-0.25em b}\kern-0.8em\TeX}}}

\begin{document}

\title{DEFT: Exploiting Gradient Norm Difference between Model Layers for Scalable Gradient Sparsification}

\author{Daegun Yoon}
\affiliation{%
  \institution{Ajou University}
  \city{Suwon}
  \country{Republic of Korea}}
\email{kljp@ajou.ac.kr}

\author{Sangyoon Oh}
\affiliation{%
  \institution{Ajou University}
  \city{Suwon}
  \country{Republic of Korea}}
\email{syoh@ajou.ac.kr}

\renewcommand{\shortauthors}{Daegun Yoon and Sangyoon Oh}

\begin{abstract}
Gradient sparsification is a widely adopted solution for reducing the excessive communication traffic in distributed deep learning. However, most existing gradient sparsifiers have relatively poor scalability because of considerable computational cost of gradient selection and/or increased communication traffic owing to gradient build-up. To address these challenges, we propose a novel gradient sparsification scheme, DEFT, that partitions the gradient selection task into sub tasks and distributes them to workers. DEFT differs from existing sparsifiers, wherein every worker selects gradients among all gradients. Consequently, the computational cost can be reduced as the number of workers increases. Moreover, gradient build-up can be eliminated because DEFT allows workers to select gradients in partitions that are non-intersecting (between workers). Therefore, even if the number of workers increases, the communication traffic can be maintained as per user requirement. 

To avoid the loss of significance of gradient selection, DEFT selects more gradients in the layers that have a larger gradient norm than the other layers. Because every layer has a different computational load, DEFT allocates layers to workers using a bin-packing algorithm to maintain a balanced load of gradient selection between workers. In our empirical evaluation, DEFT shows a significant improvement in training performance in terms of speed in gradient selection over existing sparsifiers while achieving high convergence performance.
\end{abstract}



\maketitle

\section{Introduction}\label{sec:1}
Scaling out the GPU cluster size is crucial for faster training of deep neural network (DNN) models \cite{gpt3,megatronlm,turingnlg}. However, the scalability of distributed training is affected by communication bottlenecks, where the bandwidth is insufficient to timely transmit gradients \cite{gssurvey}. To alleviate the communication problem, gradient sparsification \cite{gs01,gs02,deepgradcomp,gaussian} was introduced to reduce the number of transmitting gradients. Most of the gradient sparsification methods that have been proposed thus far \cite{adacomp,learnedcomp,errorcompensatedx,natural,sidco,accordion} share the same objective: selecting and transmitting large gradients. Because stochastic gradient descent (SGD) aims to minimise the objective function of DNN models, large gradients have a greater impact on updates than small gradients \cite{hardthreshold}. Therefore, the selection of larger gradients is preferred in gradient sparsification. Given this, Top-k gradient sparsification \cite{aji2017sparse,convproof01,convproof02,gtopk,lagssgd,omgssgd} has become the most basic approach.

However, the implementation of Top-k gradient sparsification is expensive. To identify the top $k$ elements from $n$ elements, a large amount of computational resources are required \cite{drtopk} for computational complexity of $O({n}\log{k})$ \cite{topkcomplexity}. The computational cost remains constant even if we increase the number of workers because every worker must find their top $k$ gradients from the entire gradient vector.

\begin{figure}
\centering
 \includegraphics[width=1.0\linewidth]{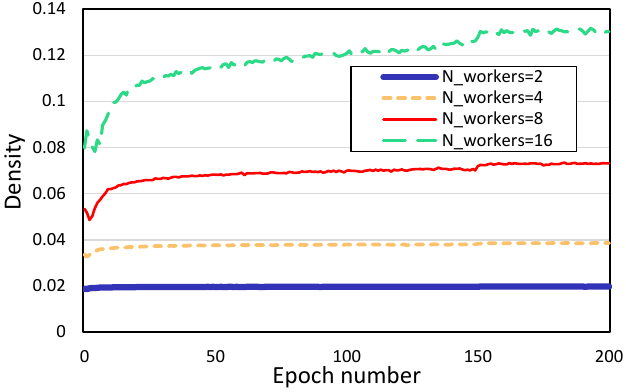}
\caption{Gradient build-up problem of Top-k gradient sparsification by scale-out of a cluster in distributed training. The measurement was conducted on ResNet-18 with CIFAR-10, and the batch size was 256 throughout. For each case, the actual density increases as the number of workers increases even though the density was set to 0.01 by the user.}
\label{fig:1}
\end{figure}

Furthermore, although every worker searches the entire gradient vector, the indices of the selected gradients by the workers do not fully overlap because every worker selects their top $k$ gradients based on their locally computed gradients (i.e., an index selected by a worker may not be selected by another worker). Therefore, to compute the average gradients, the selected indices must be collected from all the workers \cite{sketched,gradiveq,powersgd,gtopk}. Consequently, the number of globally selected indices increases, whereas the number of locally selected indices remains fixed at $k$. This phenomenon is called gradient build-up \cite{scalecom}, and may increase the communication traffic by up to $n$ times when the number of workers is $n$ in the worst case. Figure~\ref{fig:1} shows that the measurement of density\footnote{Density ($d=\frac{k}{n_g}$) is defined as the ratio of the selected gradients ($k$) to all gradients ($n_g$), and it is the inverse of compression ratio which is the term used in several papers. For example, $d=0.01$ indicates that 1$\%$ of the gradients are selected from among all gradients. In this case, the compression ratio is 100$\times$.} increases through the gradient build-up. In Figure~\ref{fig:1}, the density measured in each case was significantly higher than user-set value $0.01$, although it should be retained at $0.01$. Therefore, gradient build-up can become a negative constraint on communication in terms of scalability and must be addressed to scale-out the cluster.

Recently, Chen et al. \cite{scalecom} proposed the cyclic local top-k (CLT-k) to eliminate gradient build-up. In the proposed scheme, the top $k$ gradient selection is delegated to one worker who becomes the leader of the selection in that iteration. The leader worker then broadcasts the indices of the selected gradients to the other workers after each iteration. However, the CLT-k scheme has limitations in terms of scalability of the distributed training. Herein, while a leader worker finds the top $k$ gradients, other workers must wait for the result while being idle. As we scale-out the cluster, resource wastage increases. In addition, the computational cost of finding the top $k$ gradients cannot be reduced even if the number of workers increases; that is, the task of finding the top $k$ gradients is not divided and performed concurrently.

In this study, we propose a scheme: \textbf{d}istributed \textbf{e}xecution of \textbf{f}ragmented \textbf{t}op-k (DEFT) to address this issue. DEFT partitions the gradient vector of the entire model into multiple tensors and exclusively allocates them to workers to reduce the search space for gradient selection. The allocation of exclusive parts of the gradients to workers leads to a reduction in the computational cost of the gradient selection of each worker as the number of workers increases (i.e., parallelisation). Although it would be natural to equally divide the gradient vector by the number of workers, DEFT divides the vector into multiple tensors based on the model layers\footnote{Each tensor may be a gradient vector that corresponds to weights or biases of one layer in deep learning frameworks such as PyTorch. For simplicity, we refer to each partitioned tensor as a layer in this paper.}. Because every layer of a DNN model has a different size and gradient norm \cite{layernorm}, DEFT assigns different numbers of gradients to be selected for each layer based on the gradient norm of that layer. We claim that a sparsifier has a higher probability of selecting large gradients from layers with large gradient norms.

Upon computation of $k$ of each layer, the layers are allocated to workers based on the computational cost of each layer using the bin-packing algorithm. Consequently, the entire gradient selection load is evenly distributed among the workers. That is, the computational cost of gradient selection can be reduced as the cluster scales out. Furthermore, gradient build-up can be eliminated because workers use different parts of the gradient vector, and the collected indices are combined without overlap. Thus, the density of gradient sparsification is retained regardless of the scale-out.

This study makes the following contributions:
\begin{itemize}
    \item \textbf{Comparably high accuracy}. The accuracy of the proposed DEFT is comparable to that of non-sparsified distributed training. This achievement is mainly due to our approach, which selects large gradients from layers with large gradient norms.
    \item \textbf{Low computational cost}. DEFT incurs a significantly lower computational cost for gradient selection than the existing sparsifiers because of the load-balanced partitioning of the gradient vector.
    \item \textbf{High scalability of distributed training}. DEFT achieves a high speedup of gradient selection while retaining high accuracy when the cluster scales out.
\end{itemize}

The rest of this paper is organised as follows. Section~\ref{sec:2} presents the preliminaries for this paper. Section~\ref{sec:3} clarifies the limitations of state-of-the-art gradient sparsification methods. Section~\ref{sec:4} proposes the DEFT designed to overcome the problems stated in this paper. Section~\ref{sec:5} verifies our contributions by empirical comparisons between DEFT and existing gradient sparsifiers. Section~\ref{sec:6} concludes the paper.

\begin{algorithm}[t]
\SetAlgoVlined
\SetAlgoCaptionSeparator{}
\SetNlSty{}{}:{}
\PrintSemicolon
\SetKwInput{KwInput}{Input}
\KwInput{$G(\cdot)$: stochastic gradients\newline
Sparsify($\cdot$): gradient sparsifier\newline
$n_g$: number of gradients in model
}
\For{worker $i$ ${\in}$ $[0,n-1]$ \textnormal{\textbf{in parallel}}}{
Initialise model $x_0$ ${\in}$ ${\mathbb{R}}^{n_g}$\;
Initialise local error $e_{i,0}$ ${\leftarrow}$ $0^{n_g}$\;
\For{iteration $t$ ${\geq}$ $0$}{
${acc}_{i,t}$ ${\leftarrow}$ $e_{i,t}+{\eta}_{t}G_{i,t}(x_t)$\;
${idx}_{i,t}$ ${\leftarrow}$ Sparsify(${acc}_{i,t}$)\;
${idx}_t$ ${\leftarrow}$ AllGather(${idx}_{i,t}$)\;
$g_{i,t}$ ${\leftarrow}$ ${acc}_{i,t}[{idx}_t]$\;
$g_t$ ${\leftarrow}$ AllReduce($g_{i,t}$, SUM)\;
$x_{t+1}$ ${\leftarrow}$ $x_t-\frac{1}{n}g_t$\;
${acc}_{i,t}[{idx}_t]$ ${\leftarrow}$ $0$\;
$e_{i,t+1}$ ${\leftarrow}$ ${acc}_{i,t}$\;
}
}
\caption{Distributed SGD with error-feedback}
\label{alg:1}
\end{algorithm}

\begin{table*}
    \centering
    \caption{Strengths and weaknesses of state-of-the-art gradient sparsifiers and the proposed DEFT.}
    \label{tab:1}
    \begin{tabular}{l>\centering p{55pt}>\centering p{55pt} >\centering p{75pt} >\centering p{45pt}>\centering p{65pt} p{65pt}<\centering } 
        \toprule
        Sparsifier & Gradient build-up & Unpredictable density & Hyperparameter tuning & Worker idling & Gradient selection cost & Additional overhead \\
        \midrule
        Top-k \cite{convproof01} & Yes & Yes & No & No & Very high & No \\
        CLT-k \cite{scalecom} & No & No & No & Yes & Very high & No \\
        Hard-threshold \cite{hardthreshold} & Yes & Yes & Yes & No & Very low & No \\
        SIDCo \cite{sidco} & Yes & Yes & No & No & Very low & Very high \\
        DEFT & No & No & No & No & Low & Very low \\
        \bottomrule
    \end{tabular}
\end{table*}

\section{Preliminaries}\label{sec:2}
Gradient sparsification is in the family of lossy algorithms, because only a subset of gradients is selected while a majority of gradients are discarded. Discarding gradients leads to a difference between the trained model of gradient sparsification and that of non-sparsified distributed training. To improve the convergence rate of sparsified model, this difference must be reduced. In this section, we introduce the method that reduces the difference between sparsified- and non-sparsified- models.

We assume that a distributed SGD is scheduled at the learning rate: $\eta_{t}>0$. Subsequently, the gradient sparsification is denoted by
\begin{equation}\label{equ:1}
    x_{t+1}=x_t-{\eta_{t}}\frac{1}{n}\sum_{i=0}^{n-1}{\tilde{G}_{i,t}(x_t)},
\end{equation}
where $\tilde{G}_{i,t}(x_t)=Sparsify(G_{i,t}(x_t))$ for worker $i$ and iteration $t$. Function $Sparsify(x_t)$ selects a small fraction of the gradients from the entire gradient vector. For a gradient vector, $x{\in}n_g$, the output of $Sparsify(x_t){\in}n_g$ is a very sparse vector because the unselected gradients become zero in the vector and are discarded.

However, discarding unselected gradients is inefficient because computing gradients using backward propagation consumes considerable computing resources during distributed training. Moreover, some layers of the DNN model have larger gradients than the rest of the layers, and a sparsifier is likely to select most of the gradients only in particular layers during most of the training iterations. Although this behaviour trains a model with a fast convergence rate, the SGD may converge at an undesired point because the sparsifier conducts a biased selection of gradients. To make unselected gradients contribute to model updates, Seide et al. \cite{seide2014} introduced error-feedback that accumulates unselected gradients in the local memory of each worker. The number of unselected gradients increases owing to accumulation over the iterations. Thus, smaller gradients, when sufficiently large for selection, can also contribute to updating the model.

The sum of each gradient accumulation for each worker is referred to as the local error. In this study, the local error is denoted by ${\lVert}e_{i,t}{\rVert}$ using the L2-norm for the magnitude of the error. Then, the error is denoted as
\begin{equation}\label{equ:2}
    {\lVert}e_{t}{\rVert}=\frac{1}{n}\sum_{i=0}^{n-1}{{\lVert}e_{i,t}{\rVert}}.
\end{equation}
The error represents the update difference between models with and without sparsification. We consider that a sparsifier conducts a significant gradient selection if it effectively reduces the error.

Algorithm~\ref{alg:1} presents the pseudocode of distributed SGD to which gradient sparsification with error-feedback is applied. In line 5, the computed gradients are accumulated to local error. In line 6, the sparsifier selects the gradients based on its policy and returns the indices of the selected gradients. From lines 7 to 9, the workers collect the globally selected indices and compute the averages of the selected gradients. From lines 10 to 12, the model is updated using the averaged gradients and the accumulated value of each selected gradient is initialised to zero. If the employed sparsifier selects large gradients effectively, the error, ${\lVert}e_{t}{\rVert}$, approaches zero as the iterations proceed.

\section{Limitations of state-of-the-art methods}\label{sec:3}
In this section, we discuss the limitations of the state-of-the-art methods. Table~\ref{tab:1} lists the strengths and weaknesses of the state-of-the-art methods.

\subsection{Sorting-based Sparsifiers}\label{sec:3.1}
To determine top $k$ gradients, the entire gradient vector must be sorted. In Table~\ref{tab:1}, Top-k and CLT-k \cite{scalecom} correspond to sparsifiers that require gradient--vector sorting. Owing to sorting, these methods suffer from extremely high computational costs, thereby limiting their practicality. Moreover, this computational cost remains the same regardless of the number of workers.

In terms of communication, Top-k incurs a gradient build-up that becomes more severe as the number of workers increases, making the actual density unpredictable. In contrast, CLT-k eliminates gradient build-up by delegating the top $k$ gradient selection to one worker. Thus, the actual density is always maintained at the user-set value. However, as a side effect, CLT-k causes other workers to remain idle, which is not the case for Top-k. Therefore, Top-k and CLT-k have limitations in terms of scalability.

\subsection{Threshold-based Sparsifiers}\label{sec:3.2}
To overcome the high computational cost of sorting, threshold-based sparsifiers \cite{hardthreshold, sidco} have been proposed; hard-threshold sparsifier and SIDCo are few examples of this type. Threshold-based sparsifiers require only $O(n_g)$ for gradient selection, because they check whether each gradient is larger than the threshold. Thus, the computational cost of the gradient selection in the hard-threshold sparsifier and SIDCo is considerably low. The difference between a hard-threshold sparsifier and SIDCo lies in the determination of the threshold. For hard-threshold sparsifier, the threshold should be defined before training and should not change during training. Therefore, strict hyperparameter tuning is required for each model and dataset. In contrast, SIDCo determines the threshold based on a statistical analysis of the gradient distribution at every iteration. Thus, the threshold changes according to the gradients of the model. However, SIDCo incurs a very high computational overhead owing to the statistical analysis at every iteration.

Despite the threshold estimation of each sparsifier, there is a large difference between the number of locally selected gradients of each worker and the number that the user wants to select. This is because of inaccurate threshold estimation. An inaccurate threshold estimation limits the practicality of threshold-based sparsifiers. Moreover, they exhibit gradient build-up because each worker selects different gradient indices. Therefore, it is difficult to predict the actual density of threshold-based sparsifiers. Consequently, they must be further studied to satisfy the fundamental purpose of gradient sparsification.

The limitations of the state-of-the-art methods reveal that it is very challenging to satisfy all the criteria in Table~\ref{tab:1}. In this study, we address this challenge using a novel sparsifier with negligible overhead.

\begin{figure}
\centering
 \includegraphics[width=1.0\linewidth]{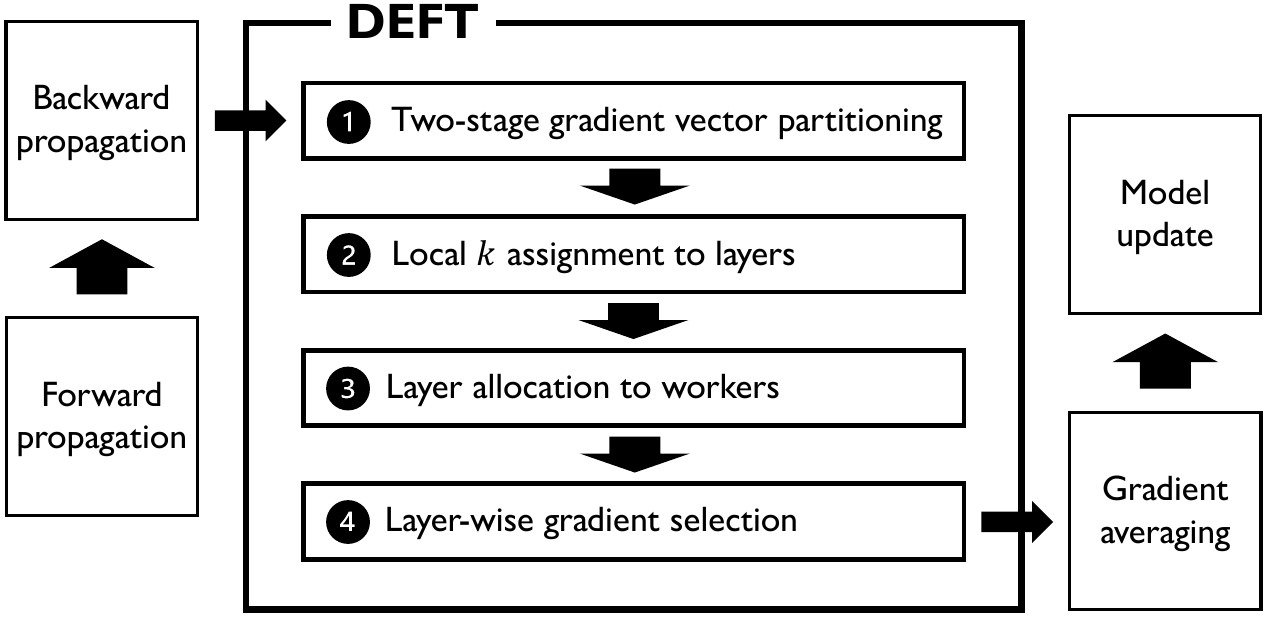}
\caption{Overview of DEFT.}
\label{fig:2}
\end{figure}

\begin{algorithm}[t]
\SetAlgoVlined
\SetAlgoCaptionSeparator{}
\SetNlSty{}{}:{}
\PrintSemicolon
\SetKwInput{KwInput}{Input}
\SetKwInput{KwOutput}{Output}
\KwInput{$g$: gradient vector\newline
$n_g$: number of gradients in model\newline
n\_workers: number of workers
}
\KwOutput{layers: list of partitioned layers
}
tensors ${\leftarrow}$ partition\_by\_layers($g$)\;
thre\_part ${\leftarrow}$ $n_g$ / n\_workers\;
layers ${\leftarrow}$ []\;
alloc\_pos ${\leftarrow}$ $0$\;
\For{tensor in tensors}{
\If{tensor.size > thre\_part}{
quotient ${\leftarrow}$ tensor.size / n\_workers\;
remainder ${\leftarrow}$ tensor.size $\%$ n\_workers\;
\For{i in range(n\_workers)}{
size\_layer ${\leftarrow}$ quotient\;
\If{remainder > $0$}{
size\_layer ${\leftarrow}$ size\_layer + $1$\;
remainder ${\leftarrow}$ remainder - $1$\;
}
st\_layer ${\leftarrow}$ alloc\_pos\;
alloc\_pos ${\leftarrow}$ alloc\_pos + size\_layer\;
end\_layer ${\leftarrow}$ alloc\_pos\;
layer ${\leftarrow}$ make\_layer(st\_layer, end\_layer)\;
layers.append(layer)\;
}
}
\Else{
st\_layer ${\leftarrow}$ alloc\_pos\;
alloc\_pos ${\leftarrow}$ alloc\_pos + tensor.size\;
end\_layer ${\leftarrow}$ alloc\_pos\;
layer ${\leftarrow}$ make\_layer(st\_layer, end\_layer)\;
layers.append(layer)\;
}
}
\caption{Two-stage gradient vector partitioning}
\label{alg:2}
\end{algorithm}

\begin{algorithm}[t]
\SetAlgoVlined
\SetAlgoCaptionSeparator{}
\SetNlSty{}{}:{}
\PrintSemicolon
\SetKwInput{KwInput}{Input}
\SetKwInput{KwOutput}{Output}
\KwInput{layers: list of layers in order of high priority\newline
$n_g$: number of gradients in model\newline
$d$: density
}
\KwOutput{layers: list of layers with local $k$
}
k\_remain ${\leftarrow}$ $n_g$ {$\cdot$} $d$\;
norm\_remain ${\leftarrow}$ $0$\;
\For{layer in layers}{
norm\_remain ${\leftarrow}$ norm\_remain + layer.norm\;
}
\For{layer in layers}{
\If{norm\_remain > $0$}{
k\_temp ${\leftarrow}$ {k\_remain} {$\cdot$} {$\frac{layer.norm}{norm\_remain}$}\;
}
\Else{
k\_temp ${\leftarrow}$ $0$\;
}
\If{layer.size < k\_temp}{
layer.k ${\leftarrow}$ layer.size\;
}
\Else{
layer.k ${\leftarrow}$ max($1$, k\_temp)\;
}
k\_remain ${\leftarrow}$ k\_remain - layer.k\;
norm\_remain ${\leftarrow}$ norm\_remain - layer.norm\;
}
\caption{Gradient norm-based local $k$ assignment}
\label{alg:3}
\end{algorithm}

\section{DEFT Design}\label{sec:4}
We designed the DEFT as a gradient sparsifier that determines gradient indices by partitioning the gradient vector. DEFT comprises the following sequence: 1) two-stage gradient vector partitioning, 2) local $k$ assignment to layers, 3) layer allocation to workers, and 4) layer-wise gradient selection. Figure~\ref{fig:2} presents an overview of the DEFT. In the process of distributed training, the DEFT is located between the backward propagation and communication for gradient averaging. The following subsections present a detailed discussion of each process in the DEFT.

\subsection{Two-Stage Gradient Vector Partitioning}\label{sec:4.1}
To reduce the computational cost of the Top-k selection in the entire gradient vector, DEFT performs two-stage partitioning. First, in DEFT, the vectors are divided into layers. Because each layer comprises a different number of parameters, some layers have a much larger number of gradients than others. Thus, DEFT equally divides a layer that has more than $\frac{n_g}{n}$ into $n$ fractions. This two-stage partitioning is required to balance the computational load between the layers. For simplicity, we refer to all partitioned fractions as layers.

Algorithm~\ref{alg:2} presents the pseudocode for the two-stage gradient vector partitioning. In line 1, in the first stage of partitioning, a gradient vector is partitioned into multiple layers. From lines 7 to 26, the second stage of partitioning is performed to partition the larger layers into smaller layers. Owing to the gradient vector partitioning, the number of gradients in each layer is less than $\frac{n_g}{n}$. Two-stage partitioning contributes to a balanced computational load between workers during layer allocation because each worker cannot take the majority of the gradients.

\subsection{Local $k$ Assignment to Layers}\label{sec:4.2}
To achieve high accuracy of a DNN model, a sparsifier should select large gradients that can provide significant updates to the model. Each layer has a different gradient norm; consequently, some layers have a larger gradient norm than others. Thus, such layers have a strong possibility of comprising large gradients. Thus, DEFT is designed to select more gradients from layers with large gradient norms. In other words, the density of each layer varies.

To apply different densities to each layer, DEFT computes the gradient norm of each layer and assigns priority to the layers in the order of the norm magnitude. Thus, high priority is assigned to the layer with a large gradient norm. Based on the given priority, DEFT executes an algorithm that assigns a local $k$ to the layers. Algorithm~\ref{alg:3} presents the pseudocode for the gradient-norm-based local $k$ assignment. The algorithm begins by initialising \textit{k\_remain} and \textit{norm\_remain} to $k$ and the sum of the gradient norms of all layers, respectively. The assignment is processed in the order of high priority. DEFT assesses the significance of each layer based on the ratio of the gradient norm to \textit{norm\_remain}. As shown in line 7, DEFT assigns $k\_remain{\cdot}\frac{layer.norm}{norm\_remain}$ as local $k$ to the layer. Whenever each layer is assigned its local $k$, \textit{k\_remain} and \textit{norm\_remain} are deducted by the assigned $k$ and the layer's gradient norm, respectively. The algorithm is terminated when all layers are assigned their local $k$.

Because the local $k$ of each layer is determined by the magnitude of its gradient norm, the size of $k$ can be large, although the number of gradients in the layer is small. Similarly, the converse is also possible. Therefore, DEFT can select larger gradients from layers with large gradient norms.

\begin{algorithm}[t]
\SetAlgoVlined
\SetAlgoCaptionSeparator{}
\SetNlSty{}{}:{}
\PrintSemicolon
\SetKwInput{KwInput}{Input}
\SetKwInput{KwOutput}{Output}
\KwInput{layers: list of layers\newline
rank: identifier of worker\newline
iteration: current iteration number\newline
n\_workers: number of workers
}
\KwOutput{alloc\_part: index list of allocated layers
}
cycle ${\leftarrow}$ iteration $\%$ n\_workers\;
curr\_part ${\leftarrow}$ (cycle + rank) $\%$ n\_workers\;
\If{rank == cycle}{
size\_bin ${\leftarrow}$ zeros(n\_workers)\;
alloc\_bin ${\leftarrow}$ empty(n\_workers)\;
cost ${\leftarrow}$ empty(layers.size)\;
\For{i in range(layers.size)}{
cost[i] ${\leftarrow}$ layers[i].size {$\cdot$} log({layers[i].k})\;
}
\For{i in range(layers.size)}{
val\_max ${\leftarrow}$ max(cost)\;
idx\_max ${\leftarrow}$ cost.index(max(cost))\;
bin\_min ${\leftarrow}$ size\_bin.index(min(size\_bin))\;
alloc\_bin[bin\_min].append(idx\_max)\;
size\_bin[bin\_min] ${\leftarrow}$ size\_bin[bin\_min] + val\_max\;
cost[idx\_max] ${\leftarrow}$ -1\;
}
}
broadcast(alloc\_bin, cycle)\;
alloc\_part ${\leftarrow}$ alloc\_bin[curr\_part]\;
\caption{Bin-packing-based layer allocation}
\label{alg:4}
\end{algorithm}

\subsection{Layer Allocation to Workers}\label{sec:4.3}
In this study, we consider the computational cost of finding the top $k$ gradients from $n_g$-sized gradient vector as ${n_g}\log{k}$ \cite{topkcomplexity}. When we expand this concept to partitioned layers, the computational cost of layer $x$ is defined as follows:
\begin{equation}\label{equ:3}
    c_x = n_{g,x}\log{k_x},
\end{equation}
where $n_{g,x}$ and $k_x$ are the number of gradients (vector size) and local $k$, respectively. Because each layer has a different local $k$ and vector size, the computational cost differs between layers. Consequently, if the same number of layers is distributed to the workers, the computational load will be unbalanced among the workers. Even if only one worker consumes a considerable amount of time for the gradient selection, the entire gradient selection process is delayed. To balance the load between workers, DEFT allocates layers to workers by considering the computational costs of the layers using a bin-packing algorithm. In DEFT, the number of items and bins are the number of layers and workers, respectively. As a bin-packing policy, the layer with the highest cost is allocated to the worker with minimum sum of costs of the layers at that time.

Algorithm~\ref{alg:4} presents the pseudocode for the bin-packing-based layer allocation. While allocating layers to workers, DEFT delegates the decision making to one worker at each iteration. Each worker becomes a decision-making worker once within every $n$ iterations in a cyclic order. The delegated worker computes computational cost $c_x$ for each layer $x$. Based on the computed computational costs, the delegated worker allocates layers to bins of workers according to the defined bin-packing policy. When all the layers are allocated to bins, the list of bins is broadcast to other workers and each worker accepts the allocated layers from its bin.

DEFT incurs an overhead from layer allocation to workers because the delegated worker requires broadcast communication to share the bins. Nevertheless, the overhead is negligible because the payload is only $4L$ bytes, where $L$ is the number of layers\footnote{$L=\sum_{i=0}^{n-1}{n_{l,i}}$, where $n_{l,i}$ is the number of layers allocated to worker $i$.}. We verify how much runtime is occupied by this overhead through experiments detailed in Section~\ref{sec:5}.

As a result of exclusive layer allocation, each worker selects gradient indices that do not overlap with those of the other workers. Although the local $k$ of each layer differs slightly between workers, the total number of selected gradients is approximately $k$ when the indices are collected. This is because all workers have the same state of the model, and the gradient computed from the same parameter does not vary significantly between workers. Therefore, DEFT can maintain the actual density as the setup density, and a gradient build-up does not occur.

\begin{algorithm}[t]
\SetAlgoVlined
\SetAlgoCaptionSeparator{}
\SetNlSty{}{}:{}
\PrintSemicolon
\SetKwInput{KwInput}{Input}
\SetKwInput{KwOutput}{Output}
\KwInput{layers: list of layers\newline
alloc\_part: index list of allocated layers
}
\KwOutput{indices: indices of gradients selected by worker\newline
$k$: number of gradients selected by worker
}
$k$ ${\leftarrow}$ $0$\;
indices ${\leftarrow}$ []\;
\For{part in alloc\_part}{
part\_idx ${\leftarrow}$ topk(layers[part].grads.abs(), layers[part].k)\;
part\_idx ${\leftarrow}$ part\_idx +layers[part].st\_layer\;
indices.extend(part\_idx)\;
$k$ ${\leftarrow}$ $k$ + layers[part].k\;
}
\caption{Layer-wise gradient selection}
\label{alg:5}
\end{algorithm}

\subsection{Layer-Wise Gradient Selection}\label{sec:4.4}
To exploit the gradient norm difference between layers, each worker executes an individual Top-k operation in each layer. This method is referred to as layer-wise gradient selection, and its pseudocode is presented in Algorithm~\ref{alg:5}. According to line 4, each worker executes only the Top-k operations in the layers allocated to them.

To discuss the computational cost of the layer-wise gradient selection of DEFT, we formulated the computational cost for worker $i$ as follows:
\begin{equation}\label{equ:4}
    C_i = \sum_{x=0}^{n_{l,i}-1}{c_x} = \sum_{x=0}^{n_{l,i}-1}{n_{g,x}\log{k_x}},
\end{equation}
where $n_{l,i}$ denotes the number of layers allocated to worker $i$. Because the runtime is determined by the worker that lastly finishes the gradient selection, the computational cost of DEFT is formulated as follows:
\begin{equation}\label{equ:5}
    C(n) = \max_{i{\in}[0,n-1]}{C_i} = \max_{i{\in}[0,n-1]}{\sum_{x=0}^{n_{l,i}-1}{n_{g,x}\log{k_x}}}.
\end{equation}

The computational cost of the Top-k sparsifier and CLT-k \cite{scalecom} for gradient selection is ${n_g}\log{k}$. Therefore, the speedup of DEFT over Top-k sparsifier and CLT-k is denoted by
\begin{equation}\label{equ:6}
    f(n)=\frac{{n_g}\log{k}}{C(n)}=\frac{{n_g}\log{k}}{\max_{i{\in}[0,n-1]}{\sum_{x=0}^{n_{l,i}-1}{n_{g,x}\log{k_x}}}}.
\end{equation}
However, the speedup, $f(n)$, is nondeterministic because the result of the layer allocation is nontrivial. Thus, for the theoretical analysis of the speedup, we consider a trivial case of layer allocation. Assume that a gradient vector is divided into $n$ partitions and each worker is allocated one partition. Then, the computational cost of the gradient selection is as follows:
\begin{equation}\label{equ:7}
    C_{trivial}(n) = \frac{n_g}{n}\log{\frac{k}{n}}.
\end{equation}
Similar to $f(n)$, the speedup of the trivial case over the Top-k sparsifier and CLT-k \cite{scalecom} is denoted by
\begin{equation}\label{equ:8}
    f_{trivial}(n) = \frac{{n_g}\log{k}}{C_{trivial}(n)} = \frac{{n_g}\log{k}}{\frac{n_g}{n}\log{\frac{k}{n}}}.
\end{equation}

It should be noted that the partitioning policy of the trivial case is more coarse-grained than that of DEFT. Thus, because $C(n){\leq}C_{trivial}(n)$ is satisfied, we also confirm that $f(n){\geq}f_{trivial}(n)$. Similarly, because $f_{trivial}(n){\geq}n$ is satisfied, we confirm that the following inequality holds:
\begin{equation}\label{equ:9}
    f(n){\geq}f_{trivial}(n){\geq}n.
\end{equation}
From (\ref{equ:9}), in terms of gradient selection, we can confirm that the theoretical speedup of DEFT over the Top-k sparsifier and CLT-k \cite{scalecom} exceeds the linear speedup. Therefore, DEFT is massively scalable owing to its outstanding speedup. We will provide the verification for (\ref{equ:9}) through experiments detailed in Section~\ref{sec:5}.

\begin{table}
    \centering
    \caption{Description of each DNN application, where $B_l$ is local batch size and $n_e$ is number of epochs.}
    \label{tab:2}
    \begin{tabular}{lllll}
        \toprule
        Application & Model & Dataset & $B_l$ & $n_e$ \\
        \midrule
        Computer vision & ResNet-18 & CIFAR-10 & $2^5$ & 200\\
        Language modelling & LSTM & WikiText-2 & $2^5$ & 90\\
        Recommendation & NCF & MovieLens-20M & $2^{16}$ & 30\\
        \bottomrule
    \end{tabular}
\end{table}

\section{Evaluation}\label{sec:5}
\subsection{Methodology}\label{sec:5.1}
\textbf{System configuration}. All experiments in this study were conducted on a cluster equipped with 32 GPUs; the number of GPUs used in each experiment was different. A cluster comprises 8 nodes, and each has 4 NVIDIA Tesla V100 GPUs, 2 20-core Intel Xeon Gold 6230 @ 2.10 GHz CPUs, and 384 GB DDR4 memory. Each worker process was executed using mpirun\footnote{In all experiments, MPI is only used for multi-process execution to automatically assign an exclusive rank to each worker.} of Open MPI 4.0.5 \cite{openmpidoc} and run on one GPU with CUDA 10.1 \cite{cudadoc}.

\textbf{Models and datasets}.
We evaluate the performance of DEFT and other sparsifiers (CLT-k and Top-k sparsifier) for three types of DNN applications: 1) computer vision via ResNet-18 \cite{resnet} on CIFAR-10 \cite{cifar}, 2) language modelling via long short-term memory (LSTM) \cite{lstm} on WikiText-2 \cite{wikitext}, and 3) recommendation system via neural collaborative filtering (NCF) \cite{ncf} on MovieLens-20M \cite{movielens20m}. Table 2 lists the descriptions of the DNN applications used in our evaluation.

\textbf{Implementation}. We implemented DEFT and other approaches on top of the deep learning framework PyTorch 1.5 \cite{torchdoc}. The communication routine for distributed training was implemented using the distributed communication package of PyTorch, and NCCL 2.4 \cite{nccl} was adopted as backend to support the multi-GPU and multi-node communication primitives optimised for NVIDIA GPUs and networking: broadcast, all-gather, and all-reduce for distributed GPU training. The source code includes everything required to reproduce the results of this study, and is available publicly at \url{https://github.com/kljp/deft}.

\begin{figure*}[t]
    \centering
    \begin{subfigure}[t]{0.331\textwidth}
        \centering
        \includegraphics[width=1.0\linewidth]{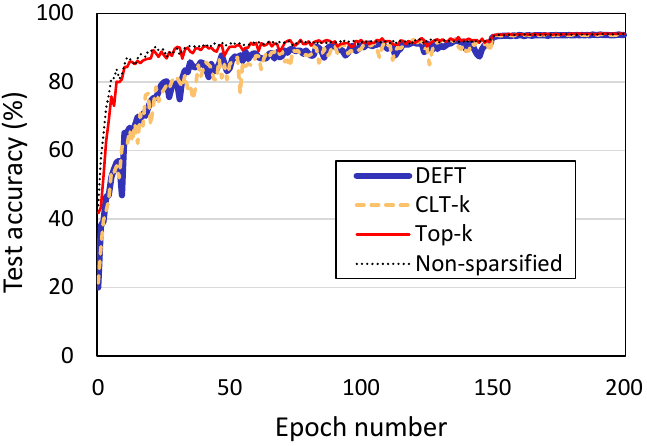}
        \caption{ResNet-18 on CIFAR-10 ($d=0.01$)}
        \label{fig:3a}
    \end{subfigure}
    ~ 
    \begin{subfigure}[t]{0.336\textwidth}
        \centering
        \includegraphics[width=1.0\linewidth]{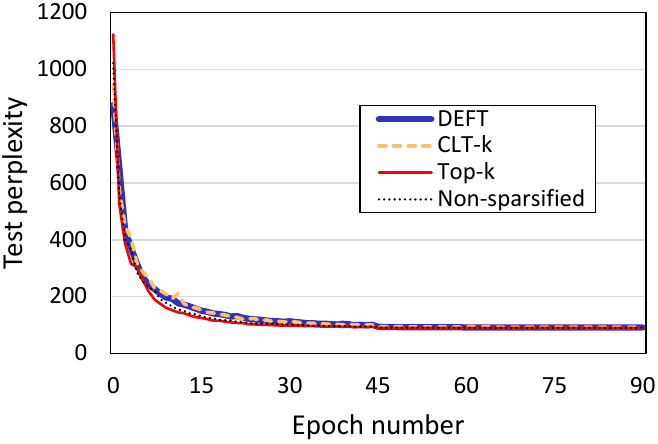}
        \caption{LSTM on WikiText-2 ($d=0.001$)}
        \label{fig:3b}
    \end{subfigure}
    ~ 
    \begin{subfigure}[t]{0.332\textwidth}
        \centering
        \includegraphics[width=1.0\linewidth]{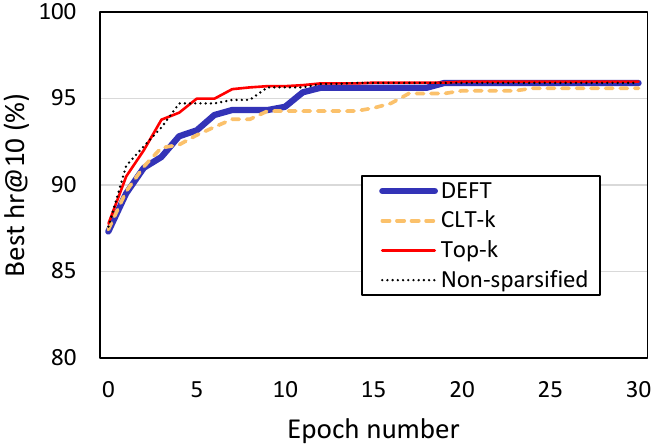}
        \caption{NCF on MovieLens-20M ($d=0.1$)}
        \label{fig:3c}
    \end{subfigure}
    \caption{Convergence performance of sparsifiers on 16 GPUs.}
    \label{fig:3}
\end{figure*}

\begin{figure*}[t]
    \centering
    \begin{subfigure}[t]{0.335\textwidth}
        \centering
        \includegraphics[width=1.0\linewidth]{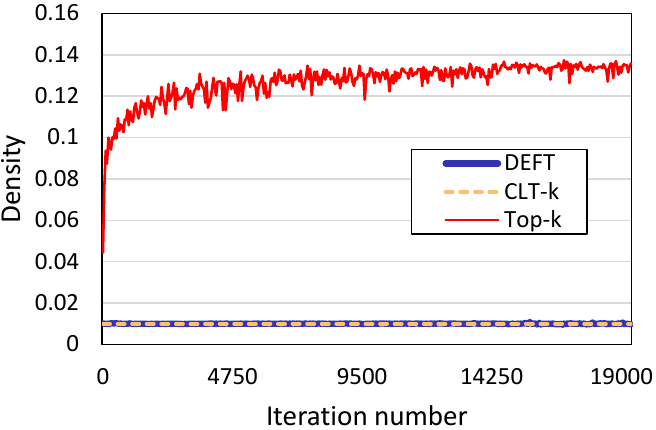}
        \caption{ResNet-18 on CIFAR-10 ($d=0.01$)}
        \label{fig:4a}
    \end{subfigure}
    ~ 
    \begin{subfigure}[t]{0.341\textwidth}
        \centering
        \includegraphics[width=1.0\linewidth]{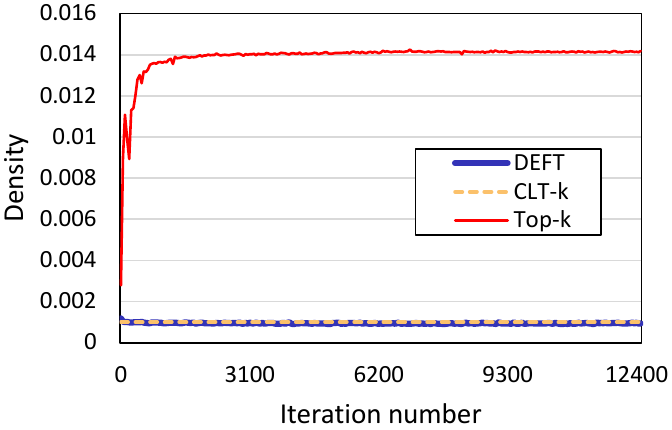}
        \caption{LSTM on WikiText-2 ($d=0.001$)}
        \label{fig:4b}
    \end{subfigure}
    ~ 
    \begin{subfigure}[t]{0.323\textwidth}
        \centering
        \includegraphics[width=1.0\linewidth]{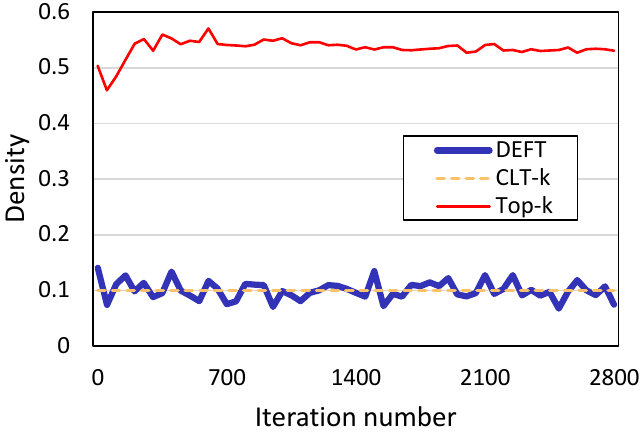}
        \caption{NCF on MovieLens-20M ($d=0.1$)}
        \label{fig:4c}
    \end{subfigure}
    \caption{Sparsification performance of sparsifiers on 16 GPUs. The Y-axis indicates the actual density measured over training iterations.}
    \label{fig:4}
\end{figure*}

\textbf{Metrics}.
The metrics used for each type of performance evaluation are as follows:
\begin{itemize}
    \item Convergence performance: Test accuracy on the computer vision application, test perplexity on language modelling application, and best Hit Rate at 10 (hr@10) \cite{htsgithub,hr10} on the recommendation system application.
    \item Sparsification performance: The actual density was measured to evaluate whether each sparsifier could maintain the density set by the user.
    \item Error minimisation performance: The error was measured to evaluate the significance of the gradient selection of each sparsifier.
\end{itemize}

\subsection{Performance Evaluation}\label{sec:5.2}
\textbf{Convergence performance}. Fig~\ref{fig:3} shows the convergence performance of each sparsifier in the DNN applications. In every application, all sparsifiers attain the convergence point of the non-sparsified distributed training, which is a distributed training without sparsification. However, the convergence rates of the sparsifiers are different. Among the three sparsifiers, the Top-k sparsifier showed the fastest convergence rate, which was close to the non-sparsified distributed training, whereas DEFT and CLT-k showed rates lower than those of the Top-k sparsifier.

\textbf{Sparsification performance}. Despite the fast convergence rate, the Top-k sparsifier showed weakness in terms of sparsification performance owing to the gradient build-up. Fig~\ref{fig:4} presents the sparsification performance of the sparsifiers on DNN applications. The Top-k sparsifier shows an excessively high actual density, whereas DEFT and CLT-k maintain the actual density at the user-set value. In this evaluation, all sparsifiers were tested on 16 GPUs, and the actual density of the Top-k sparsifier attained values of 13.59, 14.18, and 5.31 times of the setup density for the computer vision, language modelling, and recommendation system applications, respectively. In Figure~\ref{fig:4c}, the actual density of the Top-k sparsifier in the recommendation system is higher than that of the other sparsifiers, but not significantly higher as in the other two applications. This is because the Top-k sparsifier selected 53.57$\%$ of all gradients, which was sufficiently high for the workers’ gradient indices to overlap. Although DEFT showed fluctuations in the actual density in this application, it was sufficiently stable because the average density was 0.1003, which is approximate to the setup density.

Thus, it is evident that the convergence rate of the Top-k sparsifier was accomplished by a very high actual density caused by the gradient build-up. The fundamental purpose of gradient sparsification is to reduce the communication traffic. Thus, because of high actual density, Top-k sparsifier may not be an effective method despite the fast convergence rate. In contrast, DEFT can attain the convergence point of the non-sparsified distributed training at a rate similar to that of CLT-k while maintaining the actual density as the setup density.

\begin{figure*}[t]
    \centering
    \begin{subfigure}[t]{0.330\textwidth}
        \centering
        \includegraphics[width=1.0\linewidth]{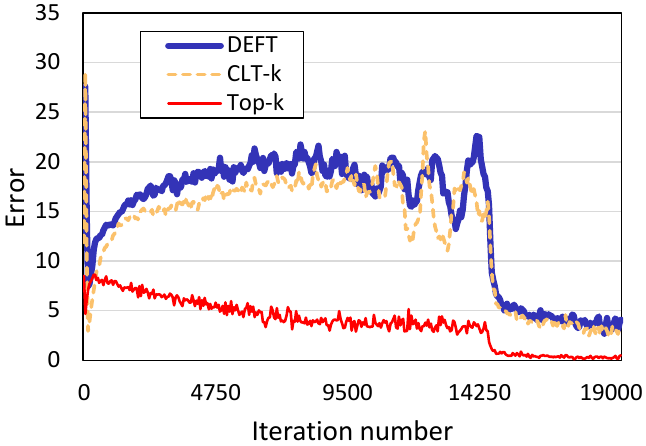}
        \caption{ResNet-18 on CIFAR-10 ($d=0.01$)}
        \label{fig:5a}
    \end{subfigure}
    ~ 
    \begin{subfigure}[t]{0.330\textwidth}
        \centering
        \includegraphics[width=1.0\linewidth]{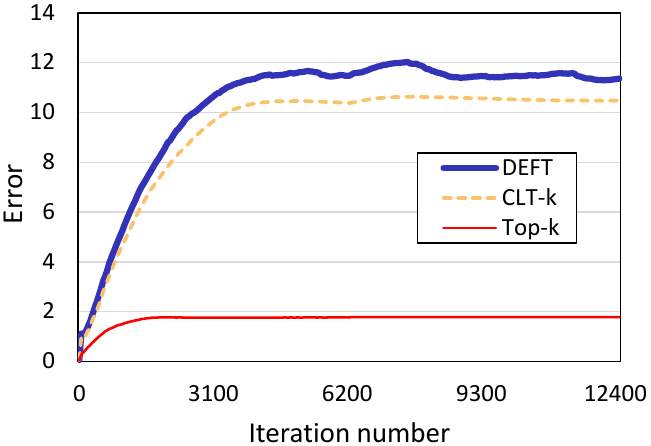}
        \caption{LSTM on WikiText-2 ($d=0.001$)}
        \label{fig:5b}
    \end{subfigure}
    ~ 
    \begin{subfigure}[t]{0.339\textwidth}
        \centering
        \includegraphics[width=1.0\linewidth]{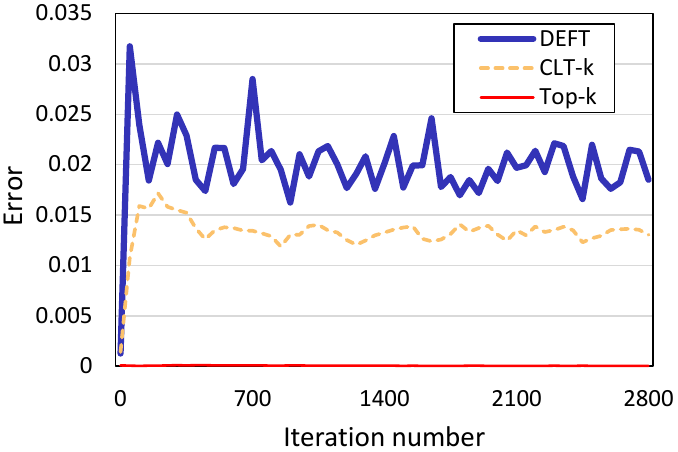}
        \caption{NCF on MovieLens-20M ($d=0.1$)}
        \label{fig:5c}
    \end{subfigure}
    \caption{Error minimisation performance of sparsifiers on 16 GPUs. The Y-axis indicates the error, which is the average of local errors of workers.}
    \label{fig:5}
\end{figure*}

\begin{figure}[t]
    \centering
    \begin{subfigure}[t]{0.232\textwidth}
        \centering
        \includegraphics[width=1.0\linewidth]{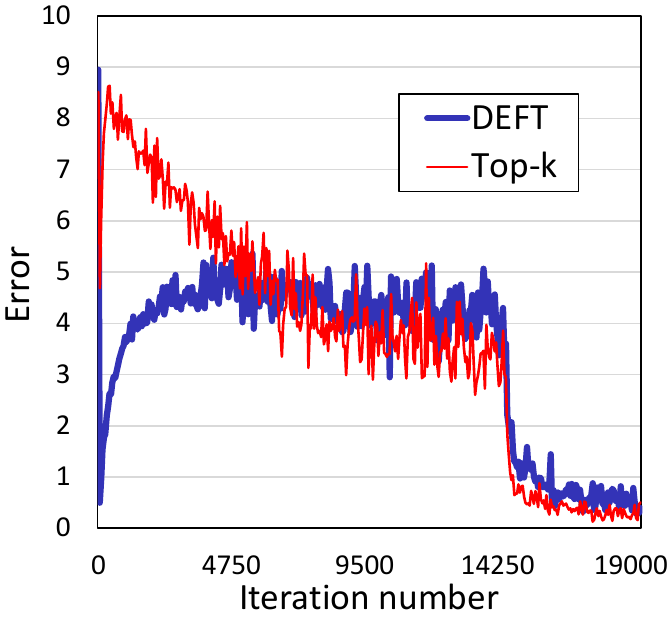}
        \caption{ResNet-18 on CIFAR-10}
        \label{fig:6a}
    \end{subfigure}
    ~ 
    \begin{subfigure}[t]{0.238\textwidth}
        \centering
        \includegraphics[width=1.0\linewidth]{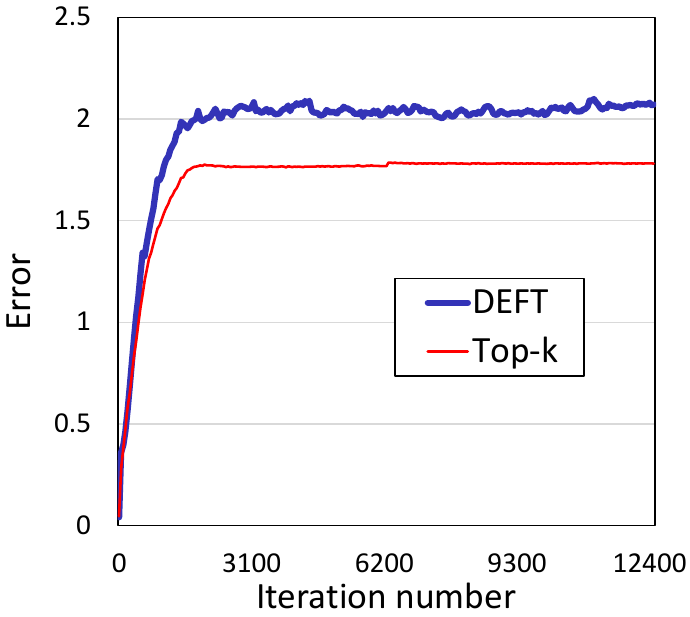}
        \caption{LSTM on WikiText-2}
        \label{fig:6b}
    \end{subfigure}
    \caption{Comparison of error minimisation performance between DEFT and Top-k sparsifier on 16 GPUs. The density of DEFT was set to a level similar to the actual density of the Top-k sparsifier. The density of DEFT was set to (a) $d=0.1$ and (b) $d=0.01$.}
    \label{fig:6}
\end{figure}

\textbf{Error Minimisation Performance}.
Figure~\ref{fig:5} presents the error-minimisation performance of the sparsifiers in DNN applications. Owing to the gradient build-up, the Top-k sparsifier shows a considerably low error compared to DEFT and CLT-k during all training iterations. This implies that the Top-k sparsifier can minimise the objective function of the DNN model faster than DEFT and CLT-k. Consequently, DEFT and CLT-k terminate with higher errors than the Top-k sparsifier at the end of model training. In Figure~\ref{fig:5a}, the sudden decrease in error after iteration 14,600 is because of learning rate decay.

As noted, the low error of the Top-k sparsifier during the training iterations is accomplished by a large number of selected gradients. To fairly compare the DEFT and Top-k sparsifiers with similar levels of actual density, we conducted additional experiments on the computer vision and language modelling applications. In these experiments, we scaled up the setup density of the DEFT to $10$ times that in the previous experiments: $d=0.1$ and $d=0.01$ for the computer vision and language modelling applications, respectively. It should be noted that these setup densities are still lower than the actual densities of the Top-k sparsifier owing to the gradient build-up. Figure~\ref{fig:6} presents the error minimisation performance between the DEFT and Top-k sparsifiers. By setting the density of the DEFT comparable to that of the Top-k sparsifier, the DEFT showed an error minimisation performance similar to that of the Top-k sparsifier.

Although DEFT selects a different number of gradients by layers, it achieves a similar level of error as the other sparsifiers. That is, DEFT can select significant gradients in the partitioned gradient vectors by considering the gradient norm of each layer, as if the CLT-k and Top-k sparsifiers find gradients in one Top-k operation in the entire gradient vector. Thus, our insight that a sparsifier should select more gradients from layers with large gradient norms is significant. Furthermore, DEFT attains the convergence point of the non-sparsified distributed training while maintaining actual density regardless of the number of workers by eliminating gradient build-up. Therefore, we conclude that DEFT has a similar level of convergence, and error minimisation performances compared to those of the other sparsifiers while achieving sparsification performance of CLT-k.

\begin{figure}
\centering
 \includegraphics[width=1.0\linewidth]{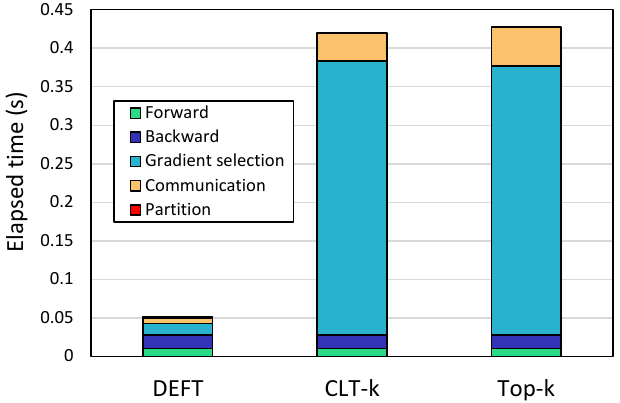}
\caption{Training time breakdown of sparsifiers on LSTM with WikiText-2. The experiments were conducted on 16 GPUs. The training time is the average wall-clock time for one iteration.}
\label{fig:7}
\end{figure}

\begin{figure}
\centering
 \includegraphics[width=1.0\linewidth]{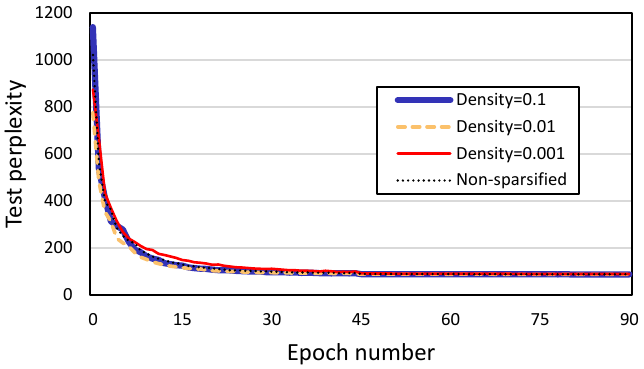}
\caption{Convergence performance of DEFT by different setup of density on LSTM with WikiText-2. The experiments were conducted on 16 GPUs.}
\label{fig:8}
\end{figure}

\subsection{Training Efficiency}\label{sec:5.3}
\textbf{Training time breakdown}. Figure~\ref{fig:7} shows a breakdown of the training time for one iteration. For each sparsifier, the wall-clock time of each iteration was measured by the slowest worker of that iteration, and the average time was calculated across all iterations. This experiment was repeated 4 times using different seeds for each sparsifier, and the average wall-clock time in Figure~\ref{fig:7} was obtained by averaging the values from 4 executions.

In Figure~\ref{fig:7}, the measured training time comprises times for forward propagation, backward propagation, gradient selection, and communication; however, DEFT also includes the partitioning overhead. DEFT requires a significantly less time for gradient selection than the CLT-k and Top-k sparsifiers. This speedup was achieved using gradient vector partitioning with a load-balanced distribution of the partitioned layers.

The communication in DEFT also reduced compared with that of the CLT-k and Top-k sparsifiers. This result can be explained by the communication time cost of the Top-k sparsifier \cite{gtopk}. The communication time cost of the Top-k sparsifier is $\log{(n)}{\alpha}+2(n-1)k{\beta}$, where $\alpha$ and $\beta$ are the system-dependent constants. In DEFT, owing to gradient partitioning, $k$ in the cost can be reduced to $\max_{i{\in}[0,n-1]}{\sum_{x=0}^{n_{l,i}-1}{k_x}}$. Accordingly, the communication time cost of DEFT is derived as $\log{(n)}{\alpha}+2(n-1)(\max_{i{\in}[0,n-1]}{\sum_{x=0}^{n_{l,i}-1}{k_x}}){\beta}$. This time cost is evidence that DEFT achieves a speedup over CLT-k and Top-k sparsifiers in communication.

Furthermore, as discussed in Section~\ref{sec:4.3}, the overhead from the layer allocation of DEFT is negligible because it is a very small fraction of the runtime in one iteration.

\textbf{Convergence performance by density}. Figure~\ref{fig:8} presents the convergence performance of DEFT for different densities. The results reveal that the convergence rate was slightly slower when the density was set to $0.001$; however, there was no significant difference after epoch 45. Thus, DEFT can achieve reasonable convergence performance with robustness to varying densities.

\begin{figure}
\centering
 \includegraphics[width=1.0\linewidth]{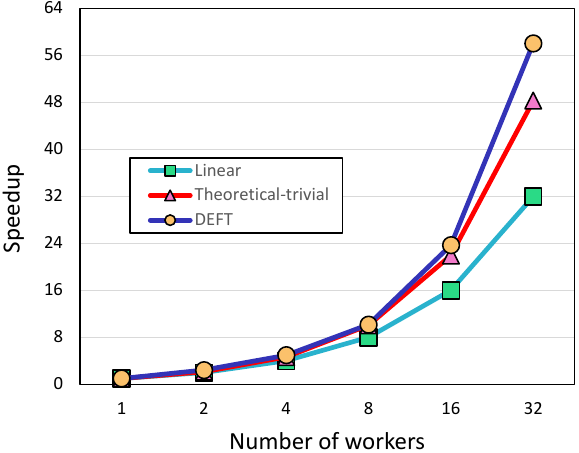}
\caption{Computational speedup of layer-wise gradient selection of DEFT over Top-k gradient selection by scale-out on LSTM with WikiText-2.}
\label{fig:9}
\end{figure}

\begin{figure}
\centering
 \includegraphics[width=1.0\linewidth]{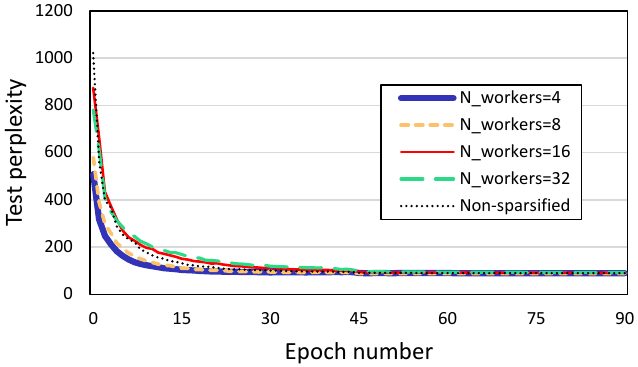}
\caption{Convergence performance of DEFT by scale-out on LSTM with WikiText-2.}
\label{fig:10}
\end{figure}

\subsection{Scalability}\label{sec:5.4}
\textbf{Computational speedup by scale}. Figure~\ref{fig:9} presents the computational speedup of the layer-wise gradient selection of DEFT over the Top-k gradient selection by scaling out the cluster. In Figure~\ref{fig:9}, if the number of workers is one, this case corresponds to the Top-k gradient selection adopted in the CLT-k and Top-k sparsifiers. The results reveal that the speedup of DEFT over the Top-k gradient selection increases as the number of workers increases.

It is noteworthy that the speedup of DEFT is higher than the linear speedup. As discussed in (\ref{equ:9}) in Section~\ref{sec:4.4}, the theoretical speedup of trivial partitioning ($f_{trivial}{(n)}$) is higher than the linear speedup. Because $f_{trivial}{(n)}$ is the worst-case speedup of DEFT, the speedup of DEFT can be higher than the linear speedup. Furthermore, it can be seen that the rate of increase of the DEFT speedup becomes significantly higher as the number of workers increases because $\frac{f_{trivial}{(n)}}{n}$ increases as $n$ increases. Therefore, the layer-wise gradient selection of DEFT has a tremendous advantage in the scale-out of the cluster.

\textbf{Convergence performance by scale}. Figure~\ref{fig:10} shows the convergence performance of the DEFT by a scale-out of the cluster. All the experiments were conducted with a setup density $0.001$. Although the convergence rate differed between cases, the test perplexity of every case eventually reached the convergence point of the non-sparsified distributed training. Because DEFT shows a high speedup over Top-k gradient selection-based sparsifiers while achieving a similar convergence rate, it is evident that the consistent convergence performance regardless of the scale of the cluster is a tremendous advantage in terms of the scalability of DEFT.

\section{Conclusion}\label{sec:6}
In this study, we proposed DEFT, which partitions the gradient vector into layers and assigns different computational loads to each layer, considering the gradient norm of the layer. The design of DEFT comprises a two-stage gradient vector partitioning, local $k$ assignment to layers, layer allocation to workers, and layer-wise gradient selection. Using these components, DEFT can achieve high performance in terms of convergence, sparsification, and error minimisation. Additionally, it provides a remarkable training efficiency owing to its reduced computational and communication costs. In particular, it exhibits a high speedup of gradient selection over the CLT-k and Top-k sparsifiers. Furthermore, it has a remarkable scalability owing to the speedup of gradient selection over linear speedup, while achieving high convergence performance.

\begin{acks}
The authors would like to thank the anonymous reviewers for their insightful feedback. This work was jointly supported by the Korea Institute of Science and Technology Information (KSC-2022-CRE-0406), BK21 FOUR program (NRF5199991014091), and Basic Science Research Program (2021R1F1A1062779) of National Research Foundation of Korea.
\end{acks}


\end{document}